\documentclass[11pt]{article}
\usepackage[utf8]{inputenc}
\usepackage[T1]{fontenc}
\usepackage[margin=1in]{geometry}
\usepackage{graphicx}
\usepackage{booktabs}
\usepackage{longtable}
\usepackage{array}
\usepackage{adjustbox}
\usepackage{pdflscape}
\usepackage{seqsplit}
\usepackage{amsmath}
\usepackage{caption}
\usepackage{hyperref}
\hypersetup{hidelinks}
\begin{document}
\begin{center}
{\LARGE Automated high-frequency quantification of fish communities and biomass using computer vision\par}
\vspace{1em}
Kota Ishikawa\textsuperscript{1,2,*} Takuma Masui\textsuperscript{3} Keita Koeda\textsuperscript{4} Rickdane Gomez\textsuperscript{3}\\
Lucas Yutaka Kimura\textsuperscript{3} Michio Kondoh\textsuperscript{1,2}\par
\vspace{0.25em}
{\setlength{\parskip}{0.2em}\small
\textsuperscript{1}Graduate School of Life Sciences, Tohoku University, Sendai, Japan\par
\textsuperscript{2}Advanced Institute for Marine Ecosystem Change (WPI-AIMEC), Tohoku University, Sendai, Japan\par
\textsuperscript{3}Graduate School of Science and Engineering, University of the Ryukyus, Okinawa, Japan\par
\textsuperscript{4}Faculty of Science, University of the Ryukyus, Okinawa, Japan\par}
\small{\textsuperscript{*}Corresponding author\par}
\vspace{1em}
\textbf{\large{Abstract}}\par
\begin{minipage}{0.9\textwidth}
Quantifying fish community structure is essential for understanding biodiversity and ecosystem responses in a changing environment, yet existing survey methods provide limited high-frequency, quantitative observations. Conventional approaches, including catch-based methods, underwater visual censuses, and environmental DNA metabarcoding, either require intensive labor or lack reliable estimates of abundance and biomass. Here, we develop an automated framework for quantifying fish communities from underwater video using computer vision. Using videos acquired with a custom-made stereo camera system, the framework integrates deep learning-based fish identification, multi-object tracking, and 3D reconstruction to estimate species-level abundance and biomass. We applied the approach to a reef fish community over a 20-day period with hourly daytime observations, revealing dynamic fluctuations in species richness, abundance, and biomass associated with changes in species composition. By comparing fish communities estimated from visual census and environmental DNA surveys, we demonstrate that our method provides complementary strengths for continuous, non-invasive, and quantitative monitoring of consistently observed species. This approach provides a scalable foundation for long-term monitoring and advances the capacity to resolve fine-scale temporal dynamics in fish communities.\par
\vspace{0.5em}
\par
\end{minipage}
\end{center}
\vspace{0.5em}
\section*{Introduction}
The urgent need to halt biodiversity loss has been globally recognized, highlighting the importance of biodiversity monitoring for conservation and ecosystem management (Convention on Biological Diversity, 2022). In marine ecosystems, fish communities play key ecological and socioeconomic roles, yet they are highly sensitive to climate change and anthropogenic disturbances (Mora et al., 2011; Pörtner and Peck, 2010). To understand current conditions and responses to conservation management, it is essential to quantify abundance, biomass, and body size because these metrics underpin ecosystem functioning, trophic interactions, and fisheries assessments (Blanchard et al., 2005; Cheung et al., 2013; Chow et al., 2026; Fontrodona-Eslava et al., 2021; Stuart-Smith et al., 2013). This highlights the importance of monitoring approaches that go beyond species presence to include quantitative estimates of fish community structure.\par
A range of methods has been developed to assess fish biodiversity, each with distinct strengths and limitations. One of the most traditional approaches is catch-based sampling, which provides reproducible and quantitative information but is labor-intensive, invasive, and inherently selective depending on operator skill and the fishing gear (reviewed in Murphy and Jenkins, 2010). Underwater visual censuses (UVCs; Brock, 1954), in which divers record fish number and size along transects, are widely used for quantitative fish community monitoring. Key limitations of UVCs are that they are labor-intensive and require trained taxonomists to dive for sampling. UVCs are also subject to observer bias in species identification and size estimation, as well as potential behavioral changes of fish caused by diver presence (reviewed in Murphy and Jenkins, 2010). To reduce observer bias and improve reproducibility, underwater video methods, such as diver-operated recordings and baited remote underwater video (BRUV) systems, have been increasingly adopted (Cappo et al., 2003; Cappo et al., 2006; Harvey et al., 2004). These methods allow reproducible and accurate estimation of fish body length using stereo cameras (Harvey and Shortis, 1995). However, baited systems can alter natural fish behavior by selectively attracting certain taxa, particularly scavengers and predators (Cappo et al., 2006; Harvey et al., 2007; Jessop et al., 2022). In addition, deployments are typically restricted to relatively short durations, usually less than an hour (Murphy and Jenkins, 2010), providing only a temporal snapshot and limited insight into daily dynamics or longer-term variability. Although environmental DNA (eDNA) metabarcoding has been used to enable monitoring with high spatiotemporal coverage (Miya et al., 2015; Thomsen et al., 2012; Valentini et al., 2016), it remains difficult to accurately quantify abundance and biomass based on the DNA read counts or copy numbers (e.g. Lamb et al., 2019). Therefore, current approaches are limited in their ability to provide high-frequency, non-invasive, and quantitative estimates of fish community structure, including abundance and biomass.\par
One potential approach to address these limitations is the application of advanced computer vision techniques. In underwater video-based methods, fish in an image are typically processed manually: individuals are detected, their species are identified, and feature points (e.g. snout and tail) are digitized for size estimation. Recent advances in computer vision techniques have enabled the automation of these processes. Convolutional neural networks (CNNs) have been among the most widely used deep learning architectures for pattern recognition (Krizhevsky et al., 2017) and form the foundation of object detection or semantic segmentation models, such as region-based CNN (Girshick et al., 2014; He et al., 2018; Ren et al., 2017) and You Only Look Once (YOLO; Redmon and Farhadi, 2018; Redmon et al., 2016). These deep learning methods are applied to classify, detect, and segment fish species in images, achieving good performance metrics on both custom-made and public training datasets  (reviewed in Al-Abri et al., 2025; Li et al., 2023). These models also allow automated body length or body mass measurements. This can be achieved by detecting key points on the fish body from segmented masks using geometric analysis, deep learning models designed for point detection, or reconstructed 3D point clouds of a fish body (reviewed in Li et al., 2023; Zhao et al., 2024). In particular, body length estimation has been extensively studied in aquaculture settings (Álvarez-Ellacuría et al., 2020; Monkman et al., 2019; Tseng et al., 2020). Despite these advances, fully automated video-based approaches for fish community monitoring remain limited, and comprehensive frameworks for estimating community metrics from videos are still lacking.\par
Here, we develop an automated framework for estimating fish communities using computer vision techniques, including deep learning-based segmentation, multi-object tracking, and 3D reconstruction. The framework was applied to reef fish communities recorded with a custom-made stereo camera at hourly intervals during daytime over a 20-day period, enabling high-frequency, quantitative assessment. We quantified abundance and biomass of each species, as well as community-level metrics, and revealed fine-scale temporal variability in fish community structure. Comparison with UVCs and eDNA metabarcoding showed that the observed strengths and limitations were consistent with expectations based on methodological characteristics. These results highlight that this approach provides a complementary and scalable foundation for high-frequency, long-term, quantitative monitoring of fish community dynamics.\par
\section*{Materials and Methods}
\subsection*{\textit{Field survey}}
Video recordings were conducted using a custom-made stereo camera system. The camera system consisted of two OAK-FFC-IMX378 cameras (Luxonis, Colorado, USA) driven by a Raspberry Pi 4 model B (Raspberry Pi Ltd, Cambridge, UK) and powered by mobile batteries (Anker 347 Power Bank, Anker Japan Co., Ltd., Tokyo, Japan). The system was enclosed in an underwater housing constructed from PVC parts. The baseline distance between the two cameras was set to 40 cm, balancing depth estimation accuracy for 3D reconstruction and field operability. A Raspberry Pi power management unit (RPZ-PowerMGR Rev2; Indoor Corgi, Ibaraki, Japan) enabled scheduled interval recording while minimizing battery consumption by placing the system in a sleeping mode when not in use.\par
The survey site was a shallow reef on the west coast of Okinawa Island, Japan (approximately 2 m deep, 26$^\circ$19'14.54"N, 127$^\circ$44'55.50"E). The camera was placed approximately 50 m from the shore. Video recordings of 10-min duration were acquired hourly during daytime (09:00--17:00), starting on the hour (e.g. 09:00--09:10) on 12--13 November 2024 and 7--27 November 2025. All videos were recorded at 4K resolution (3840$\times$2160) at 15 fps. Turbidity was measured at the same time using a logger (EPSA-CLW, JFE Advantech Co., Ltd., Hyogo, Japan). To compare fish communities across survey methods, we further conducted UVCs and eDNA metabarcoding. These surveys were conducted at approximately 10:00 on 7, 17, and 26 November 2025. Detailed methodologies are provided in the following sections.\par
\subsection*{\textit{Calibration for 3D reconstruction}}
For 3D reconstruction from 2D video data, the stereo camera was calibrated prior to field deployment. Camera calibration consisted of intrinsic and extrinsic calibration. Intrinsic calibration refers to camera-specific properties independent of camera position, such as focal length, image size, and principal point of each camera. Extrinsic calibration defines the relative positions and orientations of the two cameras (Hartley and Zisserman, 2004). The calibration was performed using the OpenCV library in Python, based on image sequences of a checkerboard moved within the field of view. To evaluate reconstruction accuracy, two markers (balls) with a known separation of 22.0 cm were moved at distances up to 8 m from the camera. Across 6,000 frames tested, the 90th percentile range of estimated distance were 20.8--23.4 cm, corresponding to a reconstruction error of approximately <10\%.\par
\subsection*{\textit{Image analysis framework}}
Fig. 1 illustrates the image analysis framework. First, we randomly sampled image frames with uniform temporal spacing, selecting 4,800 frames from the 2024 dataset and 3,743 frames from the 2025 dataset. Manual annotation was performed using CVAT software (\url{https://www.cvat.ai/}), assisted by Meta's Segment Anything Model (Kirillov et al., 2023). By taxonomic experts, fish were segmented and annotated at the finest identifiable level, including species, multi-species groups (two or three species), genus, or family. When identification was not possible at any of these levels, fish were annotated as unID (unidentified). Following standard deep learning terminology, these annotation categories are hereafter referred to as classes. Because several classes were rare (appearing in fewer than 10 frames), additional frames were sampled. Specifically, frames within ±100 frame indices of the original annotations were selected, provided they were separated by at least 10 frames. This procedure ensured a minimum of 10 annotated frames per class. In total, 8,782 frames spanning 109 classes were annotated. Information on the number of frames and instances for each class is also included in Table S3.\par
The dataset was then randomly split into a ratio of train:valid:test=6:2:2. This was done such that each class contained at least 6, 2, and 2 frames in the train, valid, and test dataset, respectively. Using this dataset, we trained two models. The first was a fish segmentation model for automated segmentation of fish without species classification. The second was a species detection model for multi-class object detection of fish taxa. For the fish segmentation model, we trained YOLO26 segmentation model (yolo26x-seg, Ultralytics) using a single ``fish'' class derived from all annotations. For the species detection model, we trained YOLO26 detection model (yolo26x, Ultralytics) on 108 classes, excluding the unID class. Both models were trained for 300 epochs with an input image size of 960 pixels.\par
Model selection across epochs was based on standard object detection metrics evaluated on the validation dataset. True positives (TP), false positives (FP), and false negatives (FN) were computed by comparing predictions with ground truth annotations. Precision, recall, and F1 score were defined as\par
\[
\mathrm{Precision}=\frac{TP}{TP+FP}, \quad \
\mathrm{Recall}=\frac{TP}{TP+FN}, \quad \
F1=2\frac{\mathrm{Precision}\times\mathrm{Recall}}{\mathrm{Precision}+\mathrm{Recall}}.
\]
Precision represents the proportion of predicted positives that are correct, while recall represents the proportion of actual positives that are correctly detected. F1 is a balanced metrics that combines precision and recall. From the precision--recall curve $P(r)$, the average precision (AP) was computed as\par
\[
AP=\int_0^1 P(r) \, dr,
\]
and mean average precision (mAP) was computed as\par
\[
mAP=\frac{\sum_{i=1}^{N} AP_i}{N},
\]
where $N$ denotes number of classes. Performance was evaluated at an intersection-over-union (IoU) threshold of 0.5 (mAP50) and across IoU thresholds from 0.5 to 0.95 (mAP50--95). Because mAP is an inclusive metrics that captures both detection accuracy and localization quality, the best model across epochs was selected based on mAP50--95. Final performance was evaluated on the held-out test dataset. For the evaluation of species detection model, to test performance variability due to dataset partitioning, we repeated the data split and training procedure for four times, and metrics were averaged across runs. The final model used for downstream analysis was selected based on a balance between F1 and mAP50.\par
After training, the models were applied to all frames to obtain fish segmentation and species bounding boxes for each frame. Inference was conducted using confidence thresholds of 0.7 for fish segmentation and 0.376 for species detection. These thresholds were selected to maximize the F1 score on the test dataset. From the fish segmentation outputs, we identified head-tail axis by applying principal component analysis (PCA) to skeletonized points of a segmentation polygon using the scikit-image package (van der Walt et al., 2014). The head-tail axis was defined by the first principal component, and head and tail positions were determined as the intersections between this axis and the boundary of the segmentation polygon. These endpoints were subsequently used to estimate body length during the 3D reconstruction. Then, we performed ByteTrack (Zhang et al., 2022) on bounding boxes derived from segmentation masks for each 10-min video clip. ByteTrack algorithm performs multi-object tracking based on associated detections between frames while incorporating low-confidence detections to recover missed objects and reduce track loss. With the tracking, each fish was assigned a unique individual ID. Class labels were then assigned to each tracked individual by combining the tracking results and species detection results. For each ID, all associated frames were collected, and species detections were matched to the tracked bounding boxes with an IoU threshold of 0.2. If species labels were available in more than 25\% of total frames, the most frequently observed label was assigned as the candidate class. If this label corresponded to a higher taxonomic level, and the individual has a finer-level class labels more than 5\% of total frames, the finer class label was adopted instead. These thresholds were empirically determined to balance label stability and sensitivity to rare detections. The above processing was performed independently for the left and right cameras of the stereo system.\par
To evaluate the performance of the image analysis pipeline, we randomly selected 35$\times$15-sec video clips from 2025 dataset, ensuring that no frames overlapped with the training dataset. Manual annotations of these clips identified 34 classes in this dataset. Tracking performance was evaluated using the TrackEval package (Luiten et al., 2021), based on the widely used metrics: multiple object tracking accuracy (MOTA), ID F1 score (IDF1), and higher order tracking accuracy (HOTA). MOTA is defined as\par
\[
\mathrm{MOTA}=1-\frac{\sum_t (FN_t+FP_t+IDSW_t)}{\sum_t GT_t}
\]
where $FN_t$, $FP_t$, $IDSW_t$, and $GT_t$ denote the numbers of false negatives, false positives, identity switches, and ground-truth objects at time $t$, respectively. IDF1 measures the accuracy of identity preservation and is defined as\par
\[
IDF1=\frac{2 \times IDTP}{2 \times IDTP+IDFP+IDFN},
\]
where $IDTP$, $IDFP$, and $IDFN$ denote true positive, false positive, and false negative identity matches. HOTA evaluates tracking performance by jointly considering detection and association accuracy, and is defined as\par
\[
HOTA=\sqrt{DetA \cdot AssA},
\]
where DetA and AssA represent detection accuracy and association accuracy, respectively (Luiten et al., 2021). MOTA primarily reflects bounding box detection accuracies, IDF1 emphasizes identity consistency, and HOTA provides a balanced measure of detection, localization, and association quality.\par
Stereo matching between the left and right cameras was performed using the calibrated stereo geometry. First, image pairs were rectified based on the calibration results to align epipolar geometry such that corresponding points lie on the same horizontal lines. Fish associations between the two views were then established using bounding box locations. For each detection in the left camera, correspondence candidates in the right camera were identified based on the vertical positions (top, center, and bottom) of bounding boxes. Based on the stereo rectification, correspondences were constrained along epipolar lines, such that matching candidates were required to have consistent vertical positions. A tolerance threshold (±20\% of bounding-box height) was introduced to account for detection noise, partial occlusions, and segmentation inconsistencies. When multiple candidate matches were found in the right camera, identity mapping was performed by assigning each left-camera ID to the right-camera ID with the largest number of temporally overlapping frames. Through this process, IDs appearing for fewer than 15 frames (1 s) were excluded from further analysis to reduce spurious short-term detections and potential double counting. For each matched pair of ID in left and right cameras, 3D reconstruction was performed using corresponding 2D points. The center of bounding box was used to obtain 3D position of each fish, while the two endpoints of head-tail axis were used to obtain body length in 3D space.\par
\begin{figure}[!htbp]
\centering
\includegraphics[width=0.9\textwidth]{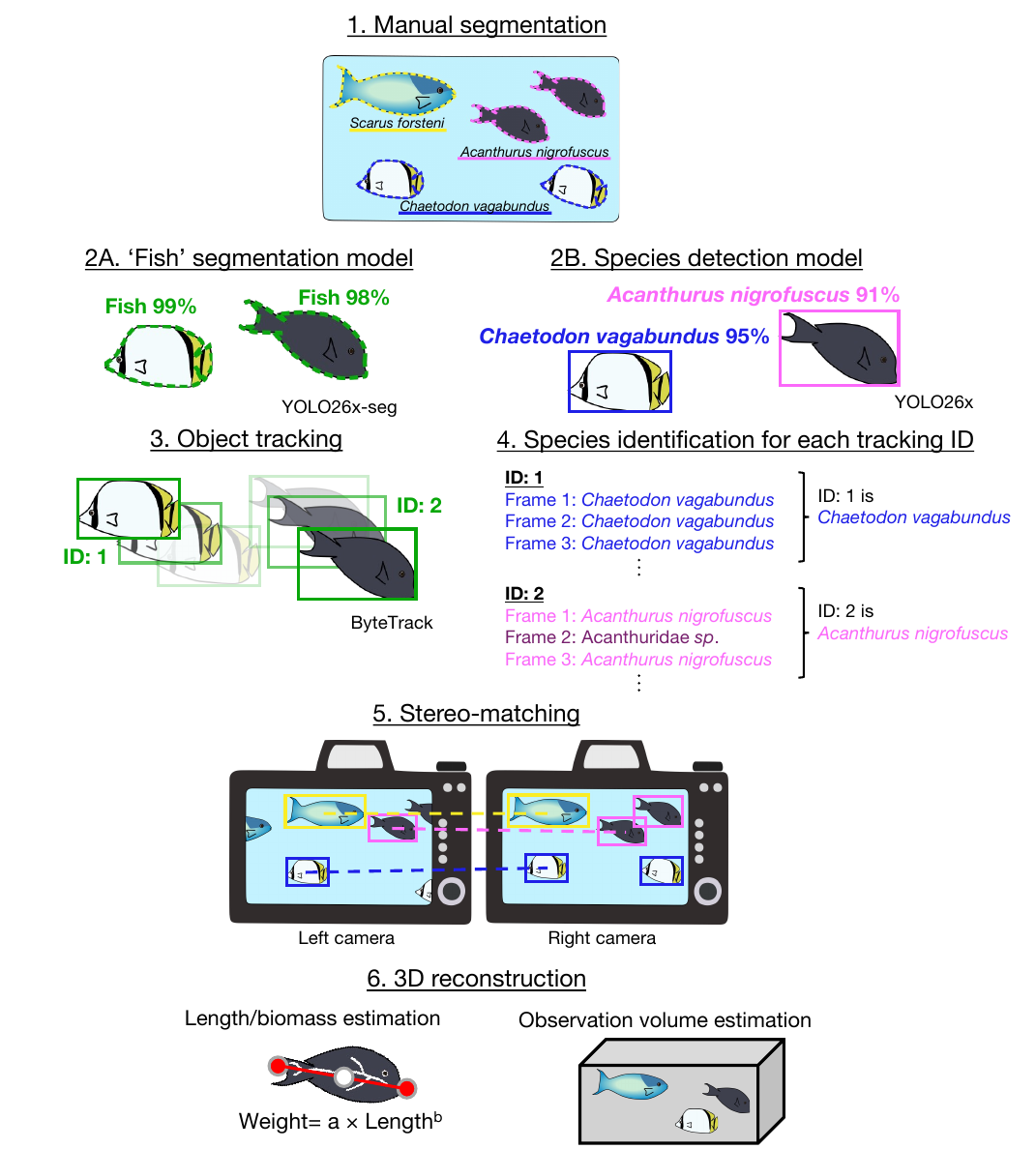}
\caption{Figure 1. Schematic overview of image analysis for fish community estimation using computer vision.}
\end{figure}
Because estimated fish body length varies across frames due to posture and orientation, length measurements were screened and aggregated for each ID. Body length estimates were retained if stereo correspondence was within the ±20\% bounding-box height threshold, either endpoint is not on the edge of image, and head--tail axis orientation is between 45$^\circ$ and 135$^\circ$ relative to the image plane. To reduce bias from orientation-dependent underestimation and exclude outliers, the 75th percentile of retained length estimate was used as the representative for each individual. Although this estimate may differ from true total length, we treated it as total length and applied species-specific length-weight relationships to convert the length to weight, following Akiona et al. (2025).\par
Based on the analysis, we computed the number of individuals and biomass of each class for each 10-min video. When computing the number of individuals, higher-level taxonomic classes were classified into species-level based on the overall ratio during the 10-min measurement. When obtaining biomass, we further excluded individuals with estimated body length exceeding 1.5 times the maximum body length registered in FishBase (Froese and Pauly, 2026) to avoid bias from extreme outliers.\par
Observation volume was estimated from 3D point clouds of detected fish aggregated over each camera deployment period (Fig. S2). A convex hull was constructed after removing outlier points, and the enclosed volume was computed as the effective observation volume. We estimated volumes during the deployment period from 9 November 11:00 to 17 November 10:00, and from 17 November 12:00 to 27 November 11:00, and observation volumes for other time periods were not estimated due to insufficient point cloud density, which prevented reliable convex hull reconstruction.\par
\subsection*{\textit{Fish community estimation by underwater visual census}}
We conducted UVCs to survey fish communities using SCUBA (Halford and Thompson, 1994). Following Gomez et al. (2025), three line transects were deployed at the survey site, each of which was 20 m long. All transects were located within a circle of 50 m radius centered on the camera. A single observer swam along each transect at an average speed of 2 m min$^{-1}$, recording all fish species observed within a transect belt extending 2.5 m on each side of the line and across the full water column (approximately 2 m depth). Each transect thus covered approximately 200 m$^3$. For each individual, total length was visually estimated to the nearest cm. Individuals smaller than 4 cm were excluded from the survey. Highly mobile fish were recorded preferentially, followed by more sedentary and site-attached species. Cryptic species were recorded when observed, including cryptobenthic fishes as defined by Brandl et al. (2018).\par
\subsection*{\textit{Fish community estimation by eDNA metabarcoding}}
For eDNA metabarcoding-based fish community estimation, we first collected surface seawater using a 1-L polypropylene bottle tied with a 10-m polyethylene rope. Sampling was conducted from shore, approximately 50 m from the camera. To prevent potential contamination, we conducted the water sampling prior to the UVC. Using two sets of a 50-mL disposable syringe and a Sterivex filter with a pore size of 0.45 $\mu$m (Merck Millipore, MA, USA), we filtered 50$\times$2 mL of collected water for each filter cartridge. We repeated this water collection and filtering process to filter 1 L of seawater for each filter in total to minimize sampling biases. Then, we added 1 mL of RNAlater (Thermo Fisher Scientific, DE, USA) to each filter cartridge to prevent eDNA degradation and stored them in a portable cooler at -18 $^\circ$C. We made a filtration blank by filtering 500 mL of milli Q water in the same way. All the filter cartridges were then moved to a -20 $^\circ$C freezer in a laboratory until eDNA extraction.\par
The eDNA on each filter sample was extracted following the eDNA Society (2020) Environmental DNA Sampling and Experiment Manual Version 2.2 with some modifications (\url{https://ednasociety.org/manual/}). After removing the RNAlater, we put extraction solution (220 $\mu$L of PBS (-), 200 $\mu$L of Buffer AL, 20 $\mu$L of Proteinase K in DNeasy Blood and Tissue kit) and incubated at 56$^\circ$C with a gentle agitation. Then, the filter unit was placed in a conical tube with a 2.0 mL tube and centrifuged at 6,000 g for 1 min to collect the extracted DNA in the 2.0 mL tube. We added 200 $\mu$L of ethanol to the extracted DNA solution and mixed by gentle pipetting. This mixture was then moved to a DNeasy Mini spin column, and the DNA was purified according to the manufacturer's protocol. The DNA was eventually eluted with 200 $\mu$L of Buffer AE. Solutions from the two filter cartridges sampled on the same day were pooled to make 400 $\mu$L of eDNA solution for each sampling. We also made extraction blank by purifying DNA from the 1 mL of purified water filtered using unused Sterivex filter during this process.\par
For eDNA metabarcoding, we then prepared solution for the first PCR. For quantitative metabarcoding (Ushio et al., 2018), we first prepared solution with known artificial sequences of 5, 10, 20, and 40 copies $\mu$L$^{-1}$ (Table S1). Then, we made 12 $\mu$L of reaction solution consisted of 2 $\mu$L of the artificial DNA solution, 2 $\mu$L of the eDNA solution, KAPA HiFi HS Ready mix (KAPA Biosystems, MA, USA), and final concentration of 0.3 $\mu$M of Primer mix (MiFishE-F/R-v2: MiFish-U-F/R: MiFish-U2-F/R = 1:2:1; 5 $\mu$M; sequence information can be found in Table S2. The mixture was then placed for first PCR that consisted of 35 cycles of 98$^\circ$C for 20 s, 65$^\circ$C for 15 s, and 72$^\circ$C for 15 s. To minimize PCR dropouts, eight technical replicates were performed. We performed the PCR on the filtering blank and extraction blank in the same way with the artificial DNA. We also prepared a 1st PCR blank, where we performed the PCR on purified water without the artificial DNA during this process.\par
After the first PCR, the samples were pooled for each eight replicates and purified using the FastGene™ Gel/PCR Extraction Kit (NIPPON Genetics, Tokyo, Japan) following the manufacturer's protocol. To quantify PCR products, we used the High Sensitivity D1000 ScreenTape System kit and TapeStation 2200 (Agilent Technologies, Tokyo, Japan) following the manufacturer's instructions. Based on the quantification results, we performed the second PCR with 12 $\mu$L of mixture, including 150 pg of the target products, KAPA HiFi HS ReadyMix, 0.3 $\mu$M of primers with the indexed sequences, and Milli Q water. We purified, quantified, and performed the 2nd PCR on filtering and extraction blanks with the artificial DNA in the same manner. Since the 1st PCR blank does not include enough amplified DNA, we diluted them based on the mean dilution ratio for all the other samples and performed 2nd PCR on them. The second PCR consisted of 10 cycles of denaturation at 98$^\circ$C for 20 s, annealing plus extension at 72$^\circ$C for 15 s, and final insertion at 72$^\circ$C for 5 min. We also made 2nd PCR blank during this process.\par
The 2nd PCR products were then purified to extract target size products derived from fish using FastGene Gel/PCR Extraction Kit (NIPPON Genetics, Tokyo, Japan), followed by E-Gel\textsuperscript{®} iBase™ Power System with E-gel SizeSelect II (Thermo Fisher Scientific, DE, USA) following the manufacturer's protocols. The size and amount of the extracted library was then analyzed using High Sensitivity DNA Kit and 2100 Bioanalyzer (Agilent Technologies, Tokyo, Japan). Using GenNext NGS Library Quantification Kit (Toyobo, Osaka, Japan) and ABI 7900HT Fast Real Time PCR System (Thermo Fisher Scientific, DE, USA), target DNA concentration for sequencing were quantified and adjusted into 4 nM. The library was then analyzed for sequencing using NextStep 500/550 Mid Output Kit v2.5 300 Cycles and NextSeq500 System (Illumina, CA, USA).\par
The bioinformatics analysis was performed using the Claident pipeline (Tanabe and Toju, 2013). The pipeline includes several steps. Following demultiplexing, raw paired-end reads were processed to remove primer sequences and to run quality filtering. Overlapping reads were merged, dereplicated, and screened for chimeras using a combination of reference-free and reference-based methods. Amplicon sequence variants (ASVs) were then inferred from high-quality reads. The QCauto + 95\%-3NN method in Claident v0.9 was used for taxonomic assignment of ASVs. This method combines a query-centric nearest-neighbor approach with lowest common ancestor logic for conservative taxonomic assignment to reduce risks of erroneous species-level identification. A curated reference database consisting of vertebrate mitochondrial 12S rRNA sequences was used (Miya et al., 2015). To minimize noise from sequencing artifacts, rare and singleton ASVs were excluded, and only taxa confidently assigned to at least the family level were retained for downstream analyses.\par
\subsection*{\textit{Comparison of fish communities estimated by different methods}}
To compare fish communities across different methods, we listed presence of species, genus, and family, based on the survey of video-based analysis, UVCs, and eDNA metabarcoding. For consistency, comparisons were based on surveys conducted on 7, 17, and 26 November, corresponding to the UVCs and eDNA sampling dates. In addition, the full 20-day results in November 2025 were used for extended comparisons. For simplicity, the subfamily Scarinae was treated as a family in this study. For visualization, hierarchical clustering was performed using pheatmap package in R (Kolde 2025), with the average linkage based on Jaccard distances of presence-absence.\par
\section*{Results}
\subsection*{\textit{Model evaluation}}
Using our training dataset, we trained a fish segmentation model and a species detection model on the YOLO26 architecture. The segmentation model achieved an mAP50 of 0.93, indicating high accuracy in fish segmentation at an IoU threshold of 0.5 (Table 1). The species detection model achieved an mAP50 of 0.63 (Table 1, YOLO26 species detection 1). This value indicates moderate overall performance, with some errors in species classification and localization. To assess effects of training data on model performance, the dataset was randomly resampled and three additional species detection models were trained. Model performance was consistent across all four runs (Table 1), indicating robustness to variation in training data. Class-level performance is summarized in Table S3. Some classes, especially those defined at genus level or combining multiple species, showed lower performance. Examples include classes, \textit{Halichoeres marginatus} or \textit{Halichoeres melanochir}, \textit{Amblyglyphidodon curacao} or \textit{Abudefduf vaigiensis} or \textit{Abudefduf sexfasciatus}, Chaetodontidae, and Siganidae. In addition, certain species such as \textit{Stethojulis bandanensis} and \textit{Arothron nigropunctatus} were detected with lower accuracy, potentially reflecting limited training data or less distinctive visual features. To examine the relationship between number of training data for each species and model performance, mAP50 values were compared across categories of class rarity. Rare classes did not consistently exhibit lower performance, although common class tended to consistently show higher performance (Fig. S1).\par
\begin{table}[htbp]
\centering
\caption{Table 1. Performance of the YOLO fish segmentation model and species detection models.}
\small
\setlength{\tabcolsep}{6pt}
\renewcommand{\arraystretch}{1.1}
\begin{adjustbox}{max width=\textwidth}
\begin{tabular}{lccccc}
\toprule
Model & Precision & Recall & F1 & mAP50 & mAP50--95 \\
\midrule
YOLO26 fish segmentation & 0.87 & 0.90 & 0.88 & 0.93 & 0.78 \\
YOLO26 species detection 1 & 0.69 & 0.57 & 0.57 & 0.63 & 0.54 \\
YOLO26 species detection 2 & 0.63 & 0.56 & 0.53 & 0.59 & 0.52 \\
YOLO26 species detection 3 & 0.70 & 0.55 & 0.55 & 0.63 & 0.54 \\
YOLO26 species detection 4 & 0.66 & 0.56 & 0.54 & 0.61 & 0.53 \\
\bottomrule
\end{tabular}
\end{adjustbox}
\end{table}
\subsection*{\textit{Evaluation of multi-object tracking}}
After multi-object tracking was applied to identify fish individuals, species labels were subsequently assigned to each tracked individual based on spatial correspondence with species detection results. To evaluate classification and tracking performance, we computed multi-object tracking metrics using 35$\times$15-sec video clips as ground truth. The MOTA, IDF1, and HOTA scores were 0.52, 0.65, and 0.65, respectively, indicating moderate tracking performance under challenging underwater conditions with variable swimming speeds and low image contrast. Class-level performance is summarized in Table S4. For some species, including \textit{Pomacentrus philippinus} and \textit{Ctenochaetus striatus}, classification performance improved at the trajectory level relative to frame-level detection, likely due to our algorithm that assigns a representative class label to each tracked individual based on predictions across frames. Classes with lower performance generally had fewer training instances and also showed lower performance in the YOLO models. Some species, such as \textit{Abudefduf vaigiensis} and \textit{Meiacanthus kamoharai}, showed reduced performance despite relatively high detection accuracy, potentially due to the small number of individuals in the ground truth video clips.\par
\subsection*{\textit{3D reconstruction}}
Following class-level tracking in the left and right cameras, 3D reconstruction was performed for the center of the bounding box, and two endpoints along the fish body axis (snout and tail). Using the reconstructed center points of stereo-matched fish, we estimated the observation volume. The observation volume varied with camera view and observation period but was approximately 70--80 m$^3$ (Fig. S2). To evaluate effects of environmental variability on detection distance, we examined relationships between turbidity and the median distance of detected fish from the camera (Y). Median Y decreased when turbidity exceeded 1 FTU (Fig. S3), indicating reduced detection range under high turbidity, which may influence estimates of individual number and biomass. With the two endpoints along the fish body axis, we estimated body length in 3D and converted it into body mass using length-weight relationships for each individual. Box plots of length and weight for each species observed during the study period in 2025 are shown in Fig S4.\par
\subsection*{\textit{Time series of community indices}}
Based on this analysis framework, species richness (number of species), abundance (number of individuals), and total biomass were quantified as community indices, along with family-level composition, for each 10-min video clip over the 20-day in 2025. These data revealed dynamic hourly fluctuations in these indices (Fig. 2). Temporal variations in these indices were associated with changes in species composition. For example, at 14:00 on Nov 14, all three indices were high, corresponding to the presence of diverse families, including relatively large-bodied taxa such as Scaridae and Acanthuridae (Supplementary video 1). At 14:00 on Nov 17, the biomass was elevated while number of species and individuals remained relatively unchanged, coinciding with the appearance of schools of \textit{Tylosurus crocodilus} and \textit{Naso unicornis} (Supplementary video 2). At 13:00 on Nov 20, abundance increased markedly, whereas biomass and species richness remained relatively constant, corresponding to the presence of a large group of Pomacentridae species (Supplementary video 3).\par
\subsection*{\textit{Comparison with other methods}}
To compare fish assemblages estimated by our method with other approaches, we conducted UVCs and eDNA metabarcoding on three separate dates. Across the 3-day survey, a total of 38 species, 29 genera, and 15 families were detected by video observation, 82 species, 55 genera, and 26 families by UVC, and 67 species, 71 genera, and 39 families by eDNA metabarcoding. Presence-absence patterns across methods are shown in Fig. 3, and Venn diagrams are shown in Fig. S5. Several species were detected by UVC but not by videos. These included strong site-attached species, such as some Pomacentridae species and \textit{Amphiprion frenatus}, which were likely outside the camera field of view. In contrast, some species were detected only by video, including \textit{Naso brachycentron}, \textit{Caranx melampygus}, \textit{Chaetodon lunulatus}, and \textit{Scarus rubroviolaceus}, which may reflect their wide-ranging behavior and relatively low occurrence. eDNA metabarcoding detected a number of cryptobenthic species (e.g. \textit{Blenniella bilitonensis}, \textit{Enneapterygius tutuilae}, \textit{Eviota guttata}, \textit{Eviota punctulata}, \textit{Macrodontogobius wilburi}, \textit{Salarias fasciatus}) and nocturnal species (e.g. \textit{Neoniphon sammara}, \textit{Parapriacanthus ransonneti}, \textit{Pempheris schwenkii}, \textit{Sargocentron diadema}) that were not observed in the visual methods. In some cases, taxa observed in the visual methods (\textit{Ctenochaetus binotatus},\textit{Naso lituratus}, \textit{Pomacentrus philippinus}, and \textit{Abudefduf sexfasciatus}) were not resolved to species level in the eDNA metabarcoding results, likely due to the conservative taxonomic assignment of the sequence analysis pipeline that minimize misidentification.\par
\begin{figure}[!htbp]
\vspace{2em}
\centering
\includegraphics[width=0.95\textwidth]{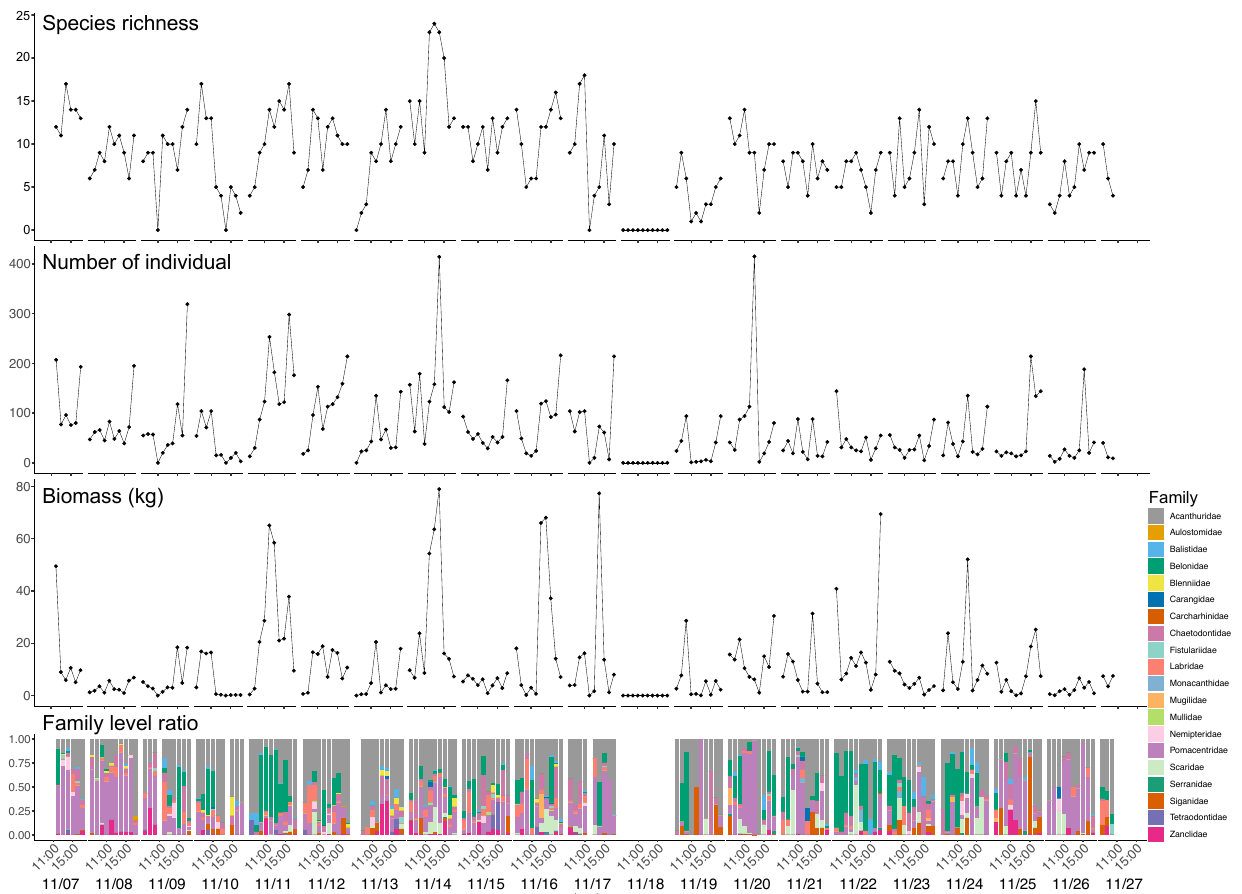}
\caption{Figure 2. Temporal variation of community indices during the 20-day survey in November 2025.}
\end{figure}

\begin{figure}[htbp]
\centering
\includegraphics[width=1\textwidth]{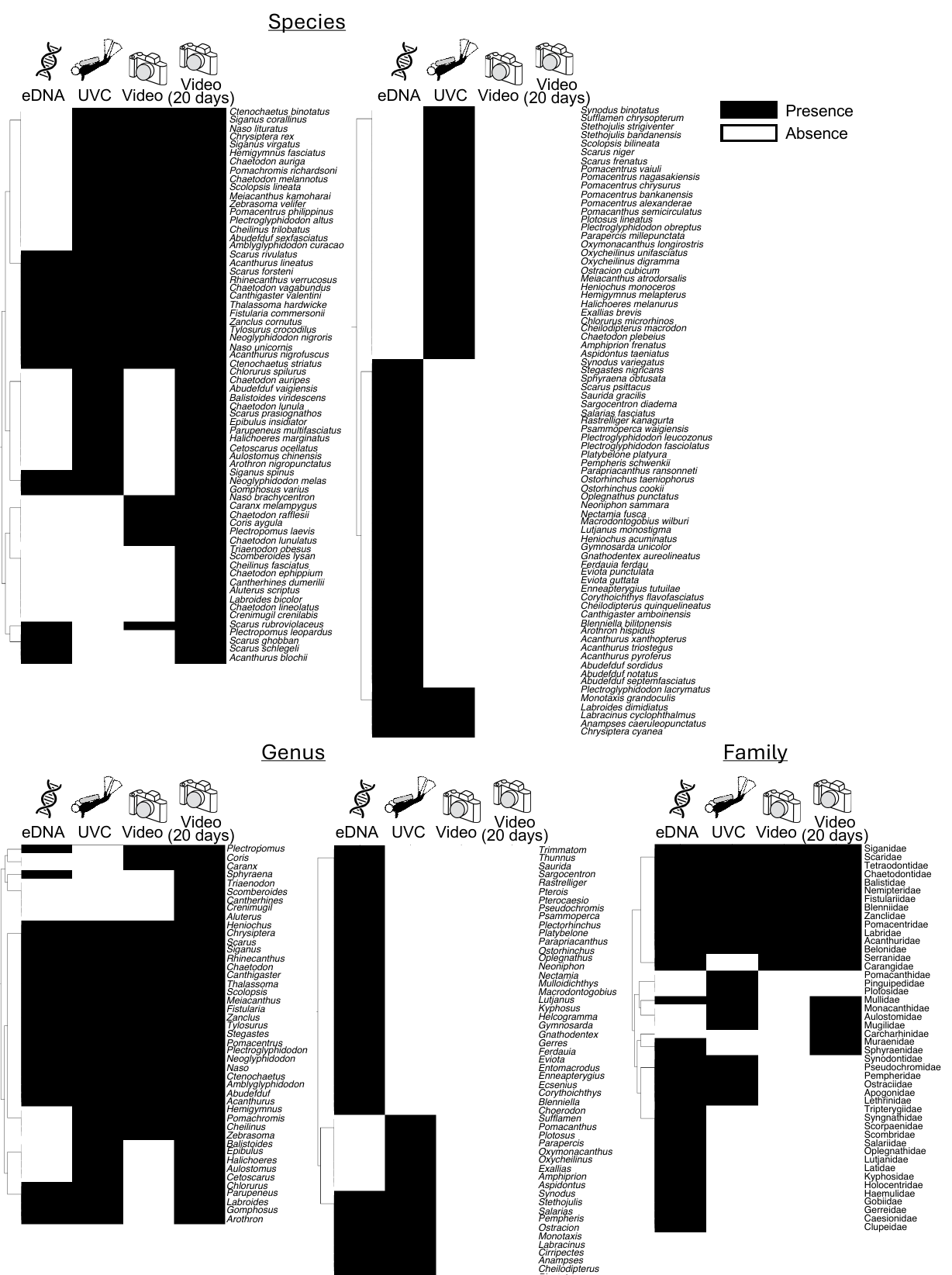}
\caption{Figure 3. Comparison of presence--absence patterns among methods.}
\end{figure}
\section*{Discussion}
We demonstrated an automated framework to estimate fish communities using computer vision. This approach quantifies community indices, including species richness, abundance, and biomass, as well as number, length, and biomass of each species or individual. With flexible recording schedules and automated processing, the framework enables high frequency and long-term monitoring with minimal human intervention. Such temporally resolved data are particularly valuable for quantitatively measuring fundamental community indices (Fontrodona-Eslava et al., 2021; Stuart-Smith et al., 2013) and accurately characterizing biodiversity dynamics in the context of ongoing biodiversity loss (Dornelas et al., 2014; Magurran et al., 2010). High-frequency and long-term time series are also critical for detecting early warning signals because indicators such as increased variance and autocorrelation emerge gradually and may be difficult to detect in short or sparsely sampled datasets (Dakos et al., 2012; Scheffer et al., 2009).\par
A major advantage of our method is its ability to provide long-term, high-frequency, and non-invasive quantification of fish communities. Although UVCs and catch-based methods can quantify abundance and biomass, eDNA metabarcoding does not reliably estimate the number of individuals or biomass for each species (Lamb et al., 2019; Rourke et al., 2022). UVCs are also subject to inter-observer biases in species identification, counts and length estimation, and fish behavior may be altered by the presence of divers (Gray et al., 2016; Harvey et al., 2004; Watson et al., 1995). Indeed, our presence-absence comparison among methods showed that some wide-ranging species (\textit{Naso brachycentron} and \textit{Caranx melampygus}) were detected by video but not by UVCs, possibly reflecting avoidance of divers. Video-based methods, such as BRUV can reduce diver-related biases but may preferentially attract scavengers and predators (Cappo et al., 2006; Harvey et al., 2007; Jessop et al., 2022). In addition, these visual methods are usually limited to short deployment periods, often less than an hour. In contrast, the proposed method minimizes observer effects and enables quantification of fish communities under more natural behavioral conditions over extended frequencies and periods.\par
From a practical perspective, our approach is also cost- and labor-efficient. The camera system used in this study cost approximately USD 1,000 and is reusable, requiring only four dives to obtain approximately 200 samples over the 20-day survey period. Once trained, the deep learning models allow fully automated video processing. In comparison, UVCs and catch-based methods would require 200 dives or sampling efforts to obtain equivalent temporal coverage. Although eDNA sampling can be automated (e.g. Hendricks et al., 2023), automated eDNA samplers are expensive, and downstream processing, including DNA extraction, PCR, library preparation, and sequencing, can cost more than USD 6,000 for 200 samples. Taken together, these comparisons highlight that our method enables scalable and high-frequency monitoring that is difficult to achieve with existing approaches.\par
One limitation of our method lies in species detection and quantification. First, the number of species detected was lower than that of other methods. This is probably because of the relatively limited observation volume, which may lead to under-detection of some cryptobenthic species (e.g. Blenniidae and Apogonidae) and highly territorial species (e.g. damselfish and anemonefish). Second, classification accuracy varied among taxa. The deep learning models showed reduced performance for some classes possibly because of the visual similarity (e.g. \textit{Halichoeres marginatus} or \textit{Halichoeres melanochir}, \textit{Amblyglyphidodon curacao} or \textit{Abudefduf vaigiensis} or \textit{Abudefduf sexfasciatus}) and indistinct visual features (e.g. \textit{Stethojulis bandanensis}, \textit{Arothron nigropunctatus}). Although higher classification accuracies have been reported in studies with fewer target species (15--23 species; Jalal et al., 2020; Salman et al., 2016; Villon et al., 2018), performance generally decreases as the number of classes increases. Given that our model included 108 classes, the observed performance (mAP50 = 0.63) is comparable to previous large-scale applications (e.g. mAP50 = 0.57 for 83 classes; Khan et al., 2023). Note that application to new regions may alter model performance and require additional model training to account for differences in species compositions and underwater conditions. Third, abundance estimates may be affected by tracking errors. Although individuals were tracked across frames, ID switching can occur when fish move rapidly or repeatedly enter and exit the field of view, potentially leading to overestimation of individual counts. These limitations could be mitigated in several ways. Increasing number of cameras would expand observation volume and improve abundance estimates and species coverage (Whitmarsh et al., 2018). Also, recent advances in computer vision, particularly transformer-based architectures, such as detection transformer (DETR) and vision transformer (ViT), may further improve classification and tracking accuracy by capturing global contextual information (Carion et al., 2020; Dosovitskiy et al., 2021). Future studies that address these limitations will further improve the reliability and applicability of automated video-based monitoring of fish communities.\par
The presence-absence comparisons across methods suggest that different methods can be used complementarily to monitor fish biodiversity, depending on research objectives. When maximizing taxonomic detection is the primary goal and quantitative estimates are not required, eDNA metabarcoding is particularly effective, as it can detect cryptobenthic and nocturnal species that are difficult to observe visually. When a balance between taxonomic coverage and quantitative monitoring is needed, UVCs, potentially combined with video methods, provide reliable identification of a wide range of species with moderate sampling frequency. When high-frequency, long-term, and minimally biased quantitative monitoring is required, our video-based framework is particularly advantageous, as it enables repeated observations of relatively representative species (i.e. species that are consistently observable within the camera field of view) under near-natural conditions.\par
Although we demonstrated this framework for quantifying fish communities, it has broader applications. By enabling 3D reconstruction, individual fish movements and behaviors can be tracked and analyzed, providing opportunities to advance ecological understanding, including inter- and intraspecific interactions (Dell et al., 2014). Time series of animal movement can also be used to classify behavioral states (Bohnslav et al., 2021; Mao et al., 2023), assess potential health conditions (Li et al., 2023), and estimate energy expenditure (Ishikawa et al., 2025a; Ishikawa et al., 2025b; Wilson et al., 2006). Furthermore, this framework may contribute to absolute fish density estimation approaches, such as random encounter staying time models (REST), which have been widely applied in terrestrial camera-trap studies and started to be extended to fish communities (Chow et al., 2026; Rowcliffe et al., 2008). Using our framework, key parameters required to build these models (e.g. staying time, movement speed of each fish) can be automatically estimated.\par
This study proposes an automated framework for quantifying fish communities from underwater video using computer vision techniques, enabling high-frequency, long-term, and non-invasive estimation of species composition, abundance, and biomass. Our observation revealed hourly fluctuations in fish communities, which were previously difficult to detect using conventional surveys. Although improvements in species coverage and classification accuracy may be needed for more reliable and broadly applicable monitoring, our framework offers a practical foundation of video-based fish community monitoring. Further studies can address these limitations by using advanced computer vision methods and deploying additional cameras, enhancing the capacity to monitor and understand dynamic fish community processes necessary for biodiversity conservation.\par
\section*{Acknowledgement}
We thank Kazusa DNA Research Institute for eDNA metabarcoding analysis and Dr. Akifumi S Tanabe for bioinformatics on the sequence data. This research was funded by the Environment Research and Technology Development Fund (JPMEERF20254RA2) of the Environmental Restoration and Conservation Agency provided by Ministry of the Environment of Japan, and the Advanced Institute for Marine Ecosystem Change (WPI-AIMEC) start-up fund awarded to KI. KK was funded by Japan Science and Technology Agency CREST (JPMJCR23J2), and KK and TM were funded by the Asahi Glass Foundation. RG was supported by the Japan Ministry of Education, Culture, Sports, Science and Technology (MEXT) Scholarship.\par
\section*{Competing interest}
All authors certify that they have no affiliations with or involvement in any organization or entity with any financial interest or non-financial interest in the subject matter or materials discussed in this manuscript.\par
\section*{Author contribution}
Conceptualization: KI; Methodology: KI; Software: KI; Formal analysis: KI; Investigation: KI, MT, KK, RG, LYK; Resources: KI; Writing - original draft: KI; Writing - review \& editing: KI, MT, KK, RG, LYK, MK; Funding acquisition: KI.\par
\section*{Data availability}
The original data and the codes are available in the GitHub repository (\url{https://github.com/Temprepo315/CVfish.git}).\par

\section*{References}
Akiona, A. K., Zgliczynski, B. J., Agarwal, M. M., French, B. J., Holloway, N. H., Lubarsky, K. A., Shirley, M. E., Sullivan, C. J. and Sandin, S. A. (2025). A database of life history parameters for Pacific coral reef fish. Sci Data 12, 1425.\par
Al-Abri, S., Keshvari, S., Al-Rashdi, K., Al-Hmouz, R. and Bourdoucen, H. (2025). Computer vision based approaches for fish monitoring systems: a comprehensive study. Artif Intell Rev 58, 185.\par
Allgeier, J. E., Layman, C. A., Mumby, P. J. and Rosemond, A. D. (2014). Consistent nutrient storage and supply mediated by diverse fish communities in coral reef ecosystems. Global Change Biology 20, 2459--2472.\par
Allgeier, J. E., Burkepile, D. E. and Layman, C. A. (2017). Animal pee in the sea: consumer-mediated nutrient dynamics in the world's changing oceans. Global Change Biology 23, 2166--2178.\par
Álvarez-Ellacuría, A., Palmer, M., Catalán, I. A. and Lisani, J.-L. (2020). Image-based, unsupervised estimation of fish size from commercial landings using deep learning. ICES J Mar Sci 77, 1330--1339.\par
Bellwood, D. R., Hughes, T. P., Folke, C. and Nyström, M. (2004). Confronting the coral reef crisis. Nature 429, 827--833.\par
Blanchard, J. L., Dulvy, N. K., Jennings, S., Ellis, J. R., Pinnegar, J. K., Tidd, A. and Kell, L. T. (2005). Do climate and fishing influence size-based indicators of Celtic Sea fish community structure? ICES J Mar Sci 62, 405--411.\par
Bohnslav, J. P., Wimalasena, N. K., Clausing, K. J., Dai, Y. Y., Yarmolinsky, D. A., Cruz, T., Kashlan, A. D., Chiappe, M. E., Orefice, L. L., Woolf, C. J., et al. (2021). DeepEthogram, a machine learning pipeline for supervised behavior classification from raw pixels. eLife 10, e63377.\par
Brandl, S. J., Goatley, C. H. R., Bellwood, D. R. and Tornabene, L. (2018). The hidden half: ecology and evolution of cryptobenthic fishes on coral reefs. Biological Reviews 93, 1846--1873.\par
Brock, V. E. (1954). A Preliminary Report on a Method of Estimating Reef Fish Populations. The Journal of Wildlife Management 18, 297--308.\par
Cappo, M., Harvey, E., Malcolm, H. and Speare, P. (2003). Potential of video techniques to monitor diversity, abundance and size of fish in studies of Marine Protected Areas. Aquatic Protected Areas-what works best and how do we know 455--464.\par
Cappo, M., Harvey, E. and Shortis, M. (2006). Counting and measuring fish with baited video techniques-an overview. AFSB CONFERENCE AND WORKSHOP" CUTTING-EDGE TECHNOLOGIES IN FISH AND FISHERIES SCIENCE 1,.\par
Carion, N., Massa, F., Synnaeve, G., Usunier, N., Kirillov, A. and Zagoruyko, S. (2020). End-to-End Object Detection with Transformers. In Computer Vision -- ECCV 2020 (ed. Vedaldi, A.), Bischof, H.), Brox, T.), and Frahm, J.-M.), pp. 213--229. Cham: Springer International Publishing.\par
Cheung, W. W. L., Sarmiento, J. L., Dunne, J., Frölicher, T. L., Lam, V. W. Y., Deng Palomares, M. L., Watson, R. and Pauly, D. (2013). Shrinking of fishes exacerbates impacts of global ocean changes on marine ecosystems. Nature Clim Change 3, 254--258.\par
Chow, C. F. Y., Brambilla, V., Fundakowski, G. J., Madin, J. S., Marques, T. A., Schiettekatte, N. M. D., Hoey, A. S. and Dornelas, M. (2026). Random encounter modelling as a viable method to estimate absolute abundance of reef fish. Methods in Ecology and Evolution 17, 598--614.\par
Convention on Biological Diversity. Kunming-Montreal Global Biodiversity Framework. Decision CBD/COP/DEC/15/4 (United Nations Environment Program, 2022).\par
Dakos, V., Carpenter, S. R., Brock, W. A., Ellison, A. M., Guttal, V., Ives, A. R., Kéfi, S., Livina, V., Seekell, D. A., Nes, E. H. van, et al. (2012). Methods for Detecting Early Warnings of Critical Transitions in Time Series Illustrated Using Simulated Ecological Data. PLOS ONE 7, e41010.\par
Dell, A. I., Bender, J. A., Branson, K., Couzin, I. D., de Polavieja, G. G., Noldus, L. P. J. J., Pérez-Escudero, A., Perona, P., Straw, A. D., Wikelski, M., et al. (2014). Automated image-based tracking and its application in ecology. Trends in Ecology \& Evolution 29, 417--428.\par
Dornelas, M., Gotelli, N. J., McGill, B., Shimadzu, H., Moyes, F., Sievers, C. and Magurran, A. E. (2014). Assemblage Time Series Reveal Biodiversity Change but Not Systematic Loss. Science 344, 296--299.\par
Dosovitskiy, A., Beyer, L., Kolesnikov, A., Weissenborn, D., Zhai, X., Unterthiner, T., Dehghani, M., Minderer, M., Heigold, G., Gelly, S., et al. (2021). An Image is Worth 16x16 Words: Transformers for Image Recognition at Scale.\par
Fontrodona-Eslava, A., Deacon, A. E., Ramnarine, I. W. and Magurran, A. E. (2021). Numerical abundance and biomass reveal different temporal trends of functional diversity change in tropical fish assemblages. Journal of Fish Biology 99, 1079--1086.\par
Froese R, Pauly D (2026) FishBase. \url{https://www.fishbase.org/}. Accessed 10 Apr 2026\par
Girshick, R., Donahue, J., Darrell, T. and Malik, J. (2014). Rich Feature Hierarchies for Accurate Object Detection and Semantic Segmentation. In 2014 IEEE Conference on Computer Vision and Pattern Recognition, pp. 580--587. Columbus, OH, USA: IEEE.\par
Gomez, R., Kimura, L. Y. and Nakamura, T. (2025). Spatial patterns and intra-annual variations in subtropical reef fish communities in Okinawa Island, Japan. Regional Studies in Marine Science 85, 104168.\par
Gray, A. E., Williams, I. D., Stamoulis, K. A., Boland, R. C., Lino, K. C., Hauk, B. B., Leonard, J. C., Rooney, J. J., Asher, J. M., Jr, K. H. L., et al. (2016). Comparison of Reef Fish Survey Data Gathered by Open and Closed Circuit SCUBA Divers Reveals Differences in Areas With Higher Fishing Pressure. PLOS ONE 11, e0167724.\par
Hartley, R. and Zisserman, A. (2004). Multiple View Geometry in Computer Vision. 2nd ed. Cambridge University Press.\par
Harvey, E. and Shortis, M. (1995). A system for stereo-video measurement of sub-tidal organisms. Marine Technology Society Journal 29, 10--22.\par
Harvey, E., Fletcher, D., Shortis, M. R. and Kendrick, G. A. (2004). A comparison of underwater visual distance estimates made by scuba divers and a stereo-video system: implications for underwater visual census of reef fish abundance. Mar. Freshwater Res. 55, 573.\par
Harvey, E. S., Cappo, M., Butler, J. J., Hall, N. and Kendrick, G. A. (2007). Bait attraction affects the performance of remote underwater video stations in assessment of demersal fish community structure. Marine Ecology Progress Series 350, 245--254.\par
He, K., Gkioxari, G., Dollár, P. and Girshick, R. (2018). Mask R-CNN.\par
Hendricks, A., Mackie, C. M., Luy, E., Sonnichsen, C., Smith, J., Grundke, I., Tavasoli, M., Furlong, A., Beiko, R. G., LaRoche, J., et al. (2023). Compact and automated eDNA sampler for in situ monitoring of marine environments. Sci Rep 13, 5210.\par
Ishikawa, K., Wu, H., Mitarai, S. and Genin, A. (2025a). Use of videos to measure dynamic body acceleration as a proxy for metabolic costs in coral reef damselfish (\textit{Chromis viridis}). Journal of Experimental Biology 228, jeb249717.\par
Ishikawa, K., Wu, H., Mitarai, S. and Genin, A. (2025b). Energy costs and benefits of locomotion and feeding in site-attached damselfish. J Exp Biol 228, jeb251164.\par
Jalal, A., Salman, A., Mian, A., Shortis, M. and Shafait, F. (2020). Fish detection and species classification in underwater environments using deep learning with temporal information. Ecological Informatics 57, 101088.\par
Jessop, S. A., Saunders, B. J., Goetze, J. S. and Harvey, E. S. (2022). A comparison of underwater visual census, baited, diver operated and remotely operated stereo-video for sampling shallow water reef fishes. Estuarine, Coastal and Shelf Science 276, 108017.\par
Khan, F. F., Li, X., Temple, A. J. and Elhoseiny, M. (2023). FishNet: A Large-scale Dataset and Benchmark for Fish Recognition, Detection, and Functional Trait Prediction. In 2023 IEEE/CVF International Conference on Computer Vision (ICCV), pp. 20439--20449.\par
Kirillov, A., Mintun, E., Ravi, N., Mao, H., Rolland, C., Gustafson, L., Xiao, T., Whitehead, S., Berg, A. C., Lo, W.-Y., et al. (2023). Segment Anything. In 2023 IEEE/CVF International Conference on Computer Vision (ICCV), pp. 3992--4003. Paris, France: IEEE.\par
Kolde R (2025). pheatmap: Pretty Heatmaps. R package version 1.0.13. URL: \url{https://CRAN.R-project.org/package=pheatmap}.\par
Krizhevsky, A., Sutskever, I. and Hinton, G. E. (2017). ImageNet classification with deep convolutional neural networks. Commun. ACM 60, 84--90.\par
Lamb, P. D., Hunter, E., Pinnegar, J. K., Creer, S., Davies, R. G. and Taylor, M. I. (2019). How quantitative is metabarcoding: A meta-analytical approach. Molecular Ecology 28, 420--430.\par
Li, J., Xu, W., Deng, L., Xiao, Y., Han, Z. and Zheng, H. (2023). Deep learning for visual recognition and detection of aquatic animals: A review. Reviews in Aquaculture 15, 409--433.\par
Luiten, J., Osep, A., Dendorfer, P., Torr, P., Geiger, A., Leal-Taixé, L. and Leibe, B. (2021). HOTA: A Higher Order Metric for Evaluating Multi-object Tracking. Int J Comput Vis 129, 548--578.\par
Magurran, A. E., Baillie, S. R., Buckland, S. T., Dick, J. McP., Elston, D. A., Scott, E. M., Smith, R. I., Somerfield, P. J. and Watt, A. D. (2010). Long-term datasets in biodiversity research and monitoring: assessing change in ecological communities through time. Trends in Ecology \& Evolution 25, 574--582.\par
Mao, A., Huang, E., Wang, X. and Liu, K. (2023). Deep learning-based animal activity recognition with wearable sensors: Overview, challenges, and future directions. Computers and Electronics in Agriculture 211, 108043.\par
Miya, M., Sato, Y., Fukunaga, T., Sado, T., Poulsen, J. Y., Sato, K., Minamoto, T., Yamamoto, S., Yamanaka, H., Araki, H., et al. (2015). MiFish, a set of universal PCR primers for metabarcoding environmental DNA from fishes: detection of more than 230 subtropical marine species. R Soc Open Sci. 2, 150088.\par
Monkman, G. G., Hyder, K., Kaiser, M. J. and Vidal, F. P. (2019). Using machine vision to estimate fish length from images using regional convolutional neural networks. Methods in Ecology and Evolution 10, 2045--2056.\par
Mora, C., Aburto-Oropeza, O., Bocos, A. A., Ayotte, P. M., Banks, S., Bauman, A. G., Beger, M., Bessudo, S., Booth, D. J., Brokovich, E., et al. (2011). Global Human Footprint on the Linkage between Biodiversity and Ecosystem Functioning in Reef Fishes. PLOS Biology 9, e1000606.\par
Murphy, H. M. and Jenkins, G. P. (2010). Observational methods used in marine spatial monitoring of fishes and associated habitats: a review. Marine and Freshwater Research 61, 236--252.\par
Pace, M. L., Cole, J. J., Carpenter, S. R. and Kitchell, J. F. (1999). Trophic cascades revealed in diverse ecosystems. Trends in Ecology \& Evolution 14, 483--488.\par
Pörtner, H. O. and Peck, M. A. (2010). Climate change effects on fishes and fisheries: towards a cause-and-effect understanding. Journal of Fish Biology 77, 1745--1779.\par
Redmon, J. and Farhadi, A. (2018). YOLOv3: An Incremental Improvement.\par
Redmon, J., Divvala, S., Girshick, R. and Farhadi, A. (2016). You Only Look Once: Unified, Real-Time Object Detection. In 2016 IEEE Conference on Computer Vision and Pattern Recognition (CVPR), pp. 779--788. Las Vegas, NV, USA: IEEE.\par
Ren, S., He, K., Girshick, R. and Sun, J. (2017). Faster R-CNN: Towards Real-Time Object Detection with Region Proposal Networks. IEEE Trans. Pattern Anal. Mach. Intell. 39, 1137--1149.\par
Rourke, M. L., Fowler, A. M., Hughes, J. M., Broadhurst, M. K., DiBattista, J. D., Fielder, S., Wilkes Walburn, J. and Furlan, E. M. (2022). Environmental DNA (eDNA) as a tool for assessing fish biomass: A review of approaches and future considerations for resource surveys. Environmental DNA 4, 9--33.\par
Rowcliffe, J. M., Field, J., Turvey, S. T. and Carbone, C. (2008). Estimating animal density using camera traps without the need for individual recognition. Journal of Applied Ecology 45, 1228--1236.\par
Salman, A., Jalal, A., Shafait, F., Mian, A., Shortis, M., Seager, J. and Harvey, E. (2016). Fish species classification in unconstrained underwater environments based on deep learning. Limnology and Oceanography: Methods 14, 570--585.\par
Scheffer, M., Bascompte, J., Brock, W. A., Brovkin, V., Carpenter, S. R., Dakos, V., Held, H., van Nes, E. H., Rietkerk, M. and Sugihara, G. (2009). Early-warning signals for critical transitions. Nature 461, 53--59.\par
Shantz, A. A., Ladd, M. C., Schrack, E. and Burkepile, D. E. (2015). Fish-derived nutrient hotspots shape coral reef benthic communities. Ecological Applications 25, 2142--2152.\par
Stuart-Smith, R. D., Bates, A. E., Lefcheck, J. S., Duffy, J. E., Baker, S. C., Thomson, R. J., Stuart-Smith, J. F., Hill, N. A., Kininmonth, S. J., Airoldi, L., et al. (2013). Integrating abundance and functional traits reveals new global hotspots of fish diversity. Nature 501, 539--542.\par
Tanabe, A. S. and Toju, H. (2013). Two New Computational Methods for Universal DNA Barcoding: A Benchmark Using Barcode Sequences of Bacteria, Archaea, Animals, Fungi, and Land Plants. PLOS ONE 8, e76910.\par
Thomsen, P. F., Kielgast, J., Iversen, L. L., Møller, P. R., Rasmussen, M. and Willerslev, E. (2012). Detection of a Diverse Marine Fish Fauna Using Environmental DNA from Seawater Samples. PLOS ONE 7, e41732.\par
Tseng, C.-H., Hsieh, C.-L. and Kuo, Y.-F. (2020). Automatic measurement of the body length of harvested fish using convolutional neural networks. Biosystems Engineering 189, 36--47.\par
Ushio, M., Murakami, H., Masuda, R., Sado, T., Miya, M., Sakurai, S., Yamanaka, H., Minamoto, T. and Kondoh, M. (2018). Quantitative monitoring of multispecies fish environmental DNA using high-throughput sequencing. Metabarcoding and Metagenomics 2, e23297.\par
Valentini, A., Taberlet, P., Miaud, C., Civade, R., Herder, J., Thomsen, P. F., Bellemain, E., Besnard, A., Coissac, E., Boyer, F., et al. (2016). Next-generation monitoring of aquatic biodiversity using environmental DNA metabarcoding. Molecular Ecology 25, 929--942.\par
van der Walt, S., Schönberger, J. L., Nunez-Iglesias, J., Boulogne, F., Warner, J. D., Yager, N., Gouillart, E. and Yu, T. (2014). scikit-image: image processing in Python. PeerJ 2, e453.\par
Villon, S., Mouillot, D., Chaumont, M., Darling, E. S., Subsol, G., Claverie, T. and Villéger, S. (2018). A Deep learning method for accurate and fast identification of coral reef fishes in underwater images. Ecological Informatics 48, 238--244.\par
Watson, R. A., Carlos, G. M. and Samoilys, M. A. (1995). Bias introduced by the non-random movement of fish in visual transect surveys. Ecological Modelling 77, 205--214.\par
Whitmarsh, S. K., Huveneers, C. and Fairweather, P. G. (2018). What are we missing? Advantages of more than one viewpoint to estimate fish assemblages using baited video. R Soc Open Sci. 5, 171993.\par
Wilson, R. P., White, C. R., Quintana, F., Halsey, L. G., Liebsch, N., Martin, G. R. and Butler, P. J. (2006). Moving towards acceleration for estimates of activity‐specific metabolic rate in free‐living animals: the case of the cormorant. Journal of Animal Ecology 75, 1081--1090.\par
Zhang, Y., Sun, P., Jiang, Y., Yu, D., Weng, F., Yuan, Z., Luo, P., Liu, W. and Wang, X. (2022). ByteTrack: Multi-Object Tracking by Associating Every Detection Box.\par
Zhao, Y., Qin, H., Xu, L., Yu, H. and Chen, Y. (2024). A review of deep learning-based stereo vision techniques for phenotype feature and behavioral analysis of fish in aquaculture. Artif Intell Rev 58, 7.\par
\end{document}